\documentclass{article}

\usepackage[preprint]{neurips_2024}

\usepackage[utf8]{inputenc} 
\usepackage[T1]{fontenc}    
\usepackage{hyperref}       
\usepackage{url}            
\usepackage{booktabs}       
\usepackage{amsfonts}       
\usepackage{nicefrac}       
\usepackage{microtype}      
\usepackage{xcolor}         

\title{Multi-objective Reinforcement Learning: A Tool for Pluralistic Alignment}

\author{%
  Peter Vamplew \\
  Institute of Innovation, Science and Sustainability\\
  Federation University Australia\\
  Mt Helen, Australia \\
  \texttt{p.vamplew@federation.edu.au} \\
  \And
  Conor F. Hayes \\
  Lawrence Livermore National Laboratory\\
  Livermore, CA, USA \\  
  \texttt{hayes56@llnl.gov} \\
  \And
  Cameron Foale \\
  Institute of Innovation, Science and Sustainability\\
  Federation University Australia\\
  Mt Helen, Australia \\
  \texttt{c.foale@federation.edu.au} \\
  \And
  Richard Dazeley \\
  School of IT\\
  Deakin University\\
  Waurn Ponds, Australia \\
\texttt{richard.dazeley@deakin.edu.au}
  \And
  Hadassah Harland \\
  School of IT\\
  Deakin University\\
  Waurn Ponds, Australia \\
\texttt{hharland@deakin.edu.au}
\\
}

\defcitealias{wang2024conditioned}{Wang, K. et al. (2024)}
\defcitealias{wang2024interpretable}{Wang, H. et al. (2024)}

\begin{document}

\maketitle

\begin{abstract}
  Reinforcement learning (RL) is a valuable tool for the creation of AI systems. However it may be problematic to adequately align RL based on scalar rewards if there are multiple conflicting values or stakeholders to be considered. Over the last decade multi-objective reinforcement learning (MORL) using vector rewards has emerged as an alternative to standard, scalar RL. This paper provides an overview of the role which MORL can play in creating pluralistically-aligned AI.    
\end{abstract}

\section{Introduction}

Reinforcement learning (RL) has emerged as one of the most powerful tools for creating AI systems capable of autonomous decision-making for sequential tasks \citep{sutton2018reinforcement}. Its core mechanism of learning to maximise the expected future return derived from a scalar reward signal can allow it to reach or even exceed human levels of performance. However, this also creates a strong dependency on an accurate definition of the reward signal. If the reward is misspecified or underspecified, the behaviour learned by an RL agent may deviate significantly from what is desired \citep{taylor2016quantilizers}. Recent studies have shown that current approaches to reward specification may frequently lead to specification errors \citep{booth2023perils,knox2023reward}.

\citet{vamplew2018human} argued that framing alignment as a multi-objective problem may assist in overcoming the issues of creating aligned agents using RL. Treating each aspect of the alignment task as a separate objective within a vector reward signal may aid in producing aligned behaviour which is difficult or impossible to achieve using a scalar definition of reward \citep{vamplew2022scalar}. Recent years have seen an increasing amount of research activity applying multi-objective reinforcement learning (MORL) techniques to various aspects of alignment.

This paper starts with a brief review of MORL, highlighting its relevance to alignment, before examining the potential use of MORL methods for pluralistic alignment, including examples of prior work. Given space restrictions these examples are illustrative rather than a comprehensive review, and will focus on alignment of large language models (LLMs) as that has been one of the main areas of application so far.  

\section{A brief review of MORL}\label{sec:morl-intro}

MORL methods assume that the environment can be represented as a Multi-objective Markov Decision Process (MOMDP), which is a MDP with a vector reward function $\mathbf{R}: S \times A \times S \to \mathbb{R}^d$ with $d$ objectives. The agent's aim is to discover a policy $\pi$ which maximises the return derived from $\mathbf{R}$. However, as both $\mathbf{R}$ and the return are vectors, it is not possible to define a full-ordering over policies. In order to do this, MORL algorithms often assume the existence of a \emph{utility function} $u:\mathbb{R}^d \to \mathbb{R}$, which maps the vector value of a policy to a scalar. If $u$ is known in advance and fixed, then the agent can learn a single-policy which is optimal for $u$ \citep{hayes2022practical}. Where $u$ is unknown, subject to change, or difficult to explicitly define, the agent may instead find a set of policies which are optimal under different parameterisations of $u$. This is known as \emph{multi-policy} MORL. The final decision of which policy to execute can then be selected at run-time. If $u$ is difficult to explicitly define, this policy selection might be carried out directly by the system's stakeholders (\citet{hayes2022practical} describe this as a \emph{decision support} scenario). 

Some MORL systems assume that $u$ is a linear-weighted sum of the objective values. This is simple to implement, as the MOMDP can be mapped to an equivalent single-objective MDP \citep{roijers2013survey}. However it may fail to correctly capture the intended behaviour, so MORL methods often instead use monotonically-increasing non-linear utility functions. This introduces algorithmic complications, but more accurately represents the stakeholders' true utility. It is important to note that in this context $u$ can not be applied to the reward received on each time-step - instead it is applied to the (possibly discounted) vector returns $\mathbf{v}$ accumulated by the agent\footnote{For clarity and brevity, we are simplifying some aspects of MORL here. For a more detailed and nuanced discussion, see \citet{hayes2022practical}.}.

\begin{equation}
    \mathbf{v} = \sum_{t=0}^T{\gamma^t\mathbf{R}_t}
    \label{eq:v}
\end{equation}

\section{Applications of MORL to pluralistic alignment}\label{sec:pluralistic-morl}

\citet{sorensen2024roadmap} define three categories of benchmarks for pluralistic alignment, which can be interpreted as specifying desirable characteristics of pluralistic agents. These characteristics are 1)  the agent be multi-objective in nature (which will support value-pluralism \citep{sorensen2024value}), 2) the agent is steerable to support customisation of trade-offs between objectives, and 3) the agent is able to consider a diverse set of user preferences. In this section we explore how MORL methods can support each of these desirable characteristics, either individually or simultaneously. 

\subsection {MORL for value pluralistic alignment}
\label{sec:value-pluralistic}

This aspect of pluralistic alignment supports consideration of a diversity of values (such as personal freedom, societal harmony, economic benefits, and environmental impact). Clearly, MORL naturally supports this aspect of pluralism, as each value can be represented by a separate objective within the reward function $\mathbf{R}$.

MORL has been applied in a number of contexts to enable a system to balance multiple conflicting values. Examples include performance versus safety tradeoffs \citep{vamplew2021potential, smith2023using}, compliance with moral standards or norms \citep{rodriguez2022instilling,peschl2021moral}, or wellbeing, affordability, equity, and environmental sustainability \citep{chaput2023learning}.

In the context of finetuning LLMs, \citetalias{wang2024conditioned} present an approach that involves conditioning model weights on preference weights over the objectives, and demonstrate performance on three objectives reflecting different desirable properties of text summarisation. \citet{wu2024fine} use fine-grained human feedback to fine-tune models based on linear combinations of rewards derived from separate reward models for factual accuracy, relevance, and information completeness. Meanwhile \citetalias{wang2024interpretable} use a context-sensitive mixture-of-experts approach to tune LLM output to trade-off between 19 different objectives which capture values such as honesty, verbosity and safety, while \citet{yang2024rewards} address the values of harmfulness, helpfulness, and humour.

\subsection{MORL for Steerable pluralistic alignment}
\label{sec:steerable-pluralistic}
A key advantage of multi-policy MORL is that it is inherently customisable with respect to stakeholder preferences. \citet{sorensen2024roadmap} state that for many applications, customisation of the trade-off between objectives at run-time is a desirable characteristic (for example, to match the preferences of the current user of an AI system). Similarly, \citet{chatila2017ieee} argued for the criticality of being able to adapt to changing values or preferences at a societal level. If the stakeholder using a system changes or if the preferences of an existing stakeholder change, a new policy which is optimal with respect to their preferences can be immediately identified, without the need for re-training which would be required for a single-objective RL agent \citep{hayes2022practical}.

\citet{harland2023ai} provides an example of these benefits. Their agent learns a Pareto set of policies, and selects a policy to execute. After each action the agent observes the reaction of a human user. If it detects that the human is displeased, the agent apologises, updates its model of the human's preferences, and selects a new policy which is compatible with that model.

\cite{rame2024rewarded} argue that for very large models, it may be impractical to explicitly learn a Pareto set of policies. Instead they start from a pre-trained model, and separately fine-tune a separate copy per objective. They then use linear weight interpolation across these specialised models at run-time to produce a model which is aligned with a particular desired trade-off between the different objectives.

\subsection{MORL for Jury-Pluralistic Alignment} \label{sec:jury-pluralistic}

Jury-pluralism refers to the capacity to take into account the variations in preferences amongst a diverse set of users (or more broadly, people or groups impacted by the decisions of the AI -- we have used the term \emph{stakeholders} to encompass all of these possibilities). In order to support this in MORL, two things must be true: (1) The set of objectives and corresponding rewards must include all aspects of the problem considered relevant by any stakeholder, and (2) the choice of policy to be executed must reflect the interests of all stakeholders. 

This might be achieved by defining in advance a utility-function \emph{u} representing the interests of all parties. In practice this is likely to be very difficult, and may well simply represent the average preferences of the population, failing to adequately consider minority views (a criticism levelled at existing reinforcement learning from human feedback (RLHF) approaches by \citet{sorensen2024roadmap}). Alternatively, multi-policy MORL methods might learn a set of Pareto-optimal policies, with the choice of policy to execute made via a consultation or voting process -- this will be time-consuming, and subject to the limitations of voting systems as well studied in social-choice literature \citep{dai2024mapping}.

Sorensen et al's definition of jury-pluralism assumes that each stakeholder is mapped to a scalar reward or utility, and that the agent makes decisions so as to maximise a welfare function applied over this set of stakeholder utilities. This definition can be mapped directly onto an MORL framework by assigning an objective within the MOMDP per stakeholder, and then applying a utility function \emph{u} that finds a suitable tradeoff over all stakeholders. %

This framework requires a diverse set $SH$ of identified stakeholders $\{{sh}_1, ...,{sh}_n \}$. Each element of the reward $\mathbf{R}$ corresponds to a scalar representation of the values of a specific stakeholder (i.e., the number of objectives $d=n$). The choice of utility function $u$ should appropriately account for the desires of each stakeholder. A relevant line of research here is the body of work on \emph{fair MORL}, where various fair utility functions have been considered. Both \cite{yu2023fair} and \cite{michailidis2023fairness} use the Generalised Gini social welfare function (GGF), which sorts the utilities of stakeholders into ascending order and applies a linear weighting with weights of decreasing value (i.e. placing more emphasis on the lower-valued utilities), as shown in Equation \ref{eq:ggf}. \cite{yu2023fair} also propose the use of an extended form, the Generalised GGF\footnote{Yes, that is the Generalised Generalised Gini social welfare function!}, which allows certain stakeholders to be prioritised. Meanwhile, \cite{fan2022welfare} propose using the Nash Social Welfare function (Equation \ref{eq:nsw}), arguing it has the benefit of being invariant to the scale of the stakeholder utilities.

\begin{center}
\begin{minipage}{.5\linewidth}
\begin{equation}
    {GGF}_\mathbf{w}(\mathbf{v}) = \sum_{i=1}^d{\mathbf{w}_i\mathbf{v}_i^\uparrow}
    \label{eq:ggf}
\end{equation}
\end{minipage}%
\begin{minipage}{.5\linewidth}
\begin{equation}
    {NSW}(\mathbf{v}) = {\left(\prod_{i=1}^d{\mathbf{v}_i}\right)}^{\frac{1}{d}}
    \label{eq:nsw}
\end{equation}
\end{minipage}
\end{center}

The concepts of individual preference models and an aggregation function have been applied in the context of LLM finetuning by \citet{park2024rlhf}. They enable the specific form of aggregation function they use (probabilistic opinion pooling) by requiring users to provide feedback in the form of a vector of the probability of selection of each sample LLM output provided in response to a prompt, rather than simply indicating a single preferred response.

\subsection{MORL for Fully Pluralistic Alignment}
\label{sec:fully-pluralistic}
The approaches described above address each dimension of pluralistic alignment separately. Jury-pluralistic MORL methods (Section \ref{sec:jury-pluralistic}) balance the preferences of a diverse set of stakeholders, while value-pluralistic MORL methods (Section \ref{sec:value-pluralistic}) find trade-offs across a diverse set of values.  Meanwhile, the steerability of multi-policy methods (Section \ref{sec:steerable-pluralistic}) applies regardless of the nature of the objectives. Here we consider how MORL methods might support a steerable agent which is both value-pluralistic and jury-pluralistic.

As in Section \ref{sec:jury-pluralistic}, a fully-pluralistic agent requires an identified set of $n$ stakeholders $SH$. However rather than representing each stakeholder's desires directly via a scalar reward value, in this framework the rewards represent varied objectives (values) which may be prioritised differently by each stakeholder. Each stakeholder's preferences over those objectives are then represented by a personalised utility function $u_i$. These individual utility functions are then aggregated by a system-level utility function. For example, the GGF from Equation \ref{eq:ggf} can be extended to this more general framework as follows:

\begin{equation}
    {GGF}_\mathbf{w}(\mathbf{v}) = \sum_{i=1}^d{\mathbf{w}_i\mathbf{u}_i^\uparrow(\mathbf{v})}
    \label{eq:ggf-plural}
\end{equation}

Multi-policy MORL methods can learn a coverage set of policies representing possible trade-offs between the objectives thereby supporting fairness across the stakeholders, while also allowing for future changes in the preferences of each stakeholder. The appropriate policy for execution can then be determined at run-time, providing a high level of steerability.

While we are not aware of any existing work on fully-pluralistic MORL agents, the PRISM Alignment Project \citep{kirk2024prism} presents an important enabling step towards fully-pluralistic LLMs, by curating a dataset of human feedback suitable for the construction of the multiple reward models required for fully-pluralistic RLHF. This feedback has been gathered from a more diverse set of participants than other feedback datasets, including attempts to provide wider geographic and cultural coverage. Each feedback item is labelled with demographic and other information about the feedback provider. In addition, the feedback is fine-grained, with a rating provided relative to multiple attributes (overall values, fluency, factuality, safety, diversity, creativity, and helpfulness). The dataset also contains measures of the importance which each provider places on each attribute, which would be suitable for creating personalised utility functions $u_i$. 

\section{Challenges}

A significant issue impeding the development of pluralistic agents using MORL is the lack of suitable human feedback datasets for developing reward models. While the PRISM dataset is a major step forward, its authors acknowledge that it still lacks sufficient diversity (for example, the content is entirely in English, and feedback was gathered from online workers) \citep{kirk2024prism}. Creation of a truly representative dataset will be a major undertaking. \citetalias{wang2024interpretable} leverage multiple datasets in an attempt to address this issue. However this is complicated as the datasets do not use consistent or compatible preference attributes. 

There is a further potential technical issue associated with the approaches proposed in Section \ref{sec:pluralistic-morl}. The majority of extant MORL algorithms are efficient only for a relatively small number of objectives (most work considers only 2-4 objectives). While this might suffice for some models such as the helpfulness/harmfulness trade-off in LLMs, it may not suffice more generallly. For example, value-pluralistic alignment based on Schwartz's theory of basic values,  would require ten objectives \citep{schwartz2012overview}. Similarly in jury-pluralistic alignment, the set of stakeholders \emph{SH} may be large, and hence the MOMDP will have a correspondingly high number of objectives. Therefore there is a need for further research extending MORL approaches to tasks involving many objectives. This may require fundamentally different methods, as has been found to be the case in optimisation, where multi-objective and many-objective cases often use different algorithms \citep{fleming2005many}. The weight interpolation approach used by \cite{rame2024rewarded} is an example of the sort of algorithmic innovation which may be required in order to scale up to high-dimensional reward spaces.

\begin{ack}
This research was supported by Founder's Pledge, the Berkeley Existential Risk Institute, and the Future of Life Institute. This work was performed under the auspices of the U.S. Department of Energy by Lawrence Livermore National Laboratory under Contract DE-AC52-07NA27344.
\end{ack}


\bibliographystyle{plainnat}      
\bibliography{pluralistic}   

\end{document}